\documentclass[pdflatex,sn-mathphys-num]{sn-jnl}

\usepackage{graphicx}%
\usepackage{multirow}%
\usepackage{amsmath,amssymb,amsfonts}%
\usepackage{amsthm}%
\usepackage{mathrsfs}%
\usepackage[title]{appendix}%
\usepackage{xcolor}%
\usepackage{textcomp}%
\usepackage{manyfoot}%
\usepackage{booktabs}%
\usepackage{subcaption}
\usepackage{amsmath, amssymb}

\usepackage{algpseudocode}
\usepackage{xcolor}

\usepackage{listings}%
\usepackage{algorithm}

\theoremstyle{thmstyleone}%
\theoremstyle{thmstyletwo}%

\theoremstyle{thmstylethree}%

\begin{document}

\title[Article Title]{Evaluating Federated Vision-Language and Vision-Only Models for Privacy-Aware Breast Cancer Imaging}



\author*[1]{\fnm{Malithi} \sur{Mithsara Wanniarachchi Kankanamge}}\email{malithi.mithsara@siu.edu}

\author[1]{\fnm{Ning} \sur{Yang}}\email{nyang@siu.edu}

\author[2]{\fnm{Sree Bash} \sur{Debnath}}\email{sdebnath@luc.edu}

\author[3]{\fnm{Md Zobaer} \sur{Islam}}\email{MdZobaer.Islam@UTSouthwestern.edu}

\author[1]{\fnm{Abdur} \sur{R. Shahid}}\email{shahid@cs.siu.edu}

\affil*[1]{\orgdiv{School of Computing}, \orgname{Southern Illinois University}, \orgaddress{\street{1230 Lincoln Drive}, \city{Carbondale}, \postcode{62901}, \state{IL}, \country{USA}}}

\affil[2]{\orgdiv{Department of Radiation Oncology, Stritch School of Medicine}, \orgname{Loyola University}, \orgaddress{\street{1032 W. Sheridan Rd}, \city{Chicago}, \postcode{60660}, \state{IL}, \country{USA}}}

\affil[3]{\orgdiv{Department of Radiology}, \orgname{University of Texas Southwestern Medical Center}, \orgaddress{\city{Dallas}, \postcode{75390}, \state{TX}, \country{USA}}}



\abstract{Artificial intelligence (AI) has shown potential for breast imaging
analysis, but its development remains challenged by limited labeled datasets,
fragmented patient data across institutions, and privacy concerns that restrict
centralized model training. We investigate a privacy-aware federated learning
framework combining pretrained vision-language models (VLMs) with distributed
training and client-level differential privacy. Alongside predictive performance,
we treat measurement as a first-class concern, since small public cohorts make it
difficult to establish that reported accuracy reflects a clinically meaningful
signal. We compare general-purpose CLIP models, the biomedical domain-specific
BiomedCLIP, and an ImageNet-pretrained ViT-B/16 baseline on two public breast
imaging datasets, under federated learning with and without client-level
differential privacy, together with an ablation removing only the text pathway from
CLIP ViT-B/16. Evaluation emphasizes patient-level ROC-AUC with bootstrap 95\%
confidence intervals, alongside explicit baselines, decision-threshold sensitivity,
inter-model agreement, and communication cost. Under federated learning, all five
models achieve high discrimination on the normal-versus-sick task, with ROC-AUC
from 0.940 to 0.990; however, acquisition year alone attains 0.974 on the same test
partition, so this performance cannot be attributed to thermal pathology.
Benign-versus-malignant performance is uncertain, with ROC-AUC from 0.504 to 0.637
and all intervals including 0.50. The ablation shows that removing the text pathway
matches full-model accuracy while halving communication, and differential privacy
reduces ranking performance by 0.375 AUC, traced to perturbing the full model state
rather than the trainable subset. These findings guide the design and evaluation of
distributed breast imaging systems that jointly consider predictive reliability,
patient privacy, and deployment cost.}




\maketitle

\section{Introduction}\label{sec:intro}

Breast cancer remains a major health concern among women worldwide, and early screening plays an important role in improving treatment outcomes and reducing mortality~\cite{niell2024acr}. Medical imaging is central to this process because it provides visual evidence to identify and assess breast abnormalities. As imaging data become increasingly complex, artificial intelligence (AI) has become a valuable tool for supporting image analysis, screening, and clinical decision-making~\cite{diaz2024artificial,radak2023machine}. However, conventional deep learning methods for breast image classification typically learn a direct mapping from an input image to a predefined diagnostic label and often require large amounts of labeled training data to achieve robust performance. This requirement is particularly challenging in medical imaging, where expert annotation is costly, time-consuming, and often limited by the availability of sufficiently large and well-curated datasets~\cite{jin2026learning}.

In response to these challenges, recent research has increasingly explored vision-language models (VLMs), which learn shared representations of images and textual descriptions~\cite{radford2021learning, wang2026generalist}. Through large-scale image-text pretraining, these models can reduce reliance on task-specific labeled data and support classification in limited-data settings. By mapping visual features and textual class descriptions into a shared embedding space, Contrastive Language-Image Pre-training (CLIP)-style models~\cite{wang2022medclip, ghosh2024mammo}, including BiomedCLIP~\cite{zhang2023biomedclip}, can associate breast images with class-specific semantic representations. However, the privacy concerns associated with patient data remain. When VLMs are adapted through centralized training, sensitive medical images from participating institutions must be transferred to a central server, raising concerns about patient privacy, data governance, and unauthorized access~\cite{li2024centralized}.

In this context, federated learning (FL) offers a promising approach by enabling collaborative model training while keeping raw medical images at their sources~\cite{rauniyar2023federated, miloudi2026federated}. In a federated setting, participating clients train models locally using their own data and communicate model updates, rather than the underlying images, to an aggregation server~\cite{koutsoubis2025privacy}. The server combines these updates to refine a shared global model, enabling learning across distributed datasets without requiring the centralized collection of patient images~\cite{raza2025federated}.

However, federated learning alone is insufficient to guarantee patient privacy. Although raw medical images remain local, the model updates exchanged during training may still reveal sensitive information about local data~\cite{koutsoubis2025privacy}. Differential privacy (DP) provides an additional layer of protection by limiting the information that can be inferred from these updates. In client-level DP, clipping bounds each client’s contribution, while calibrated random noise added during aggregation limits the influence of any individual client’s data on the resulting global model~\cite{hemalatha2025review}.

Motivated by these limitations, this study investigates pretrained visual representations for privacy-preserving federated breast cancer classification. We evaluate four CLIP-style small vision--language models and an ImageNet-pretrained Vision Transformer (ViT) baseline under federated learning with and without client-level differential privacy, and complement this comparison with a controlled ablation that removes only the text pathway from CLIP ViT-B/16 while retaining its image encoder. Breast images remain distributed across simulated clients, while only model updates are communicated to an aggregation server. To address the limited size and class imbalance of the evaluated datasets, the training pipeline incorporates image augmentation, minority-class oversampling, class weighting, and focal loss, and distributes the training data across clients using a non-IID Dirichlet partition. Throughout, we evaluate at the patient level with confidence intervals and explicit baselines, and examine whether the learned signal can be attributed to the imaging content. The main contributions of this work are summarized as follows:

\begin{figure*}[t]
\centering
\includegraphics[width=\textwidth]{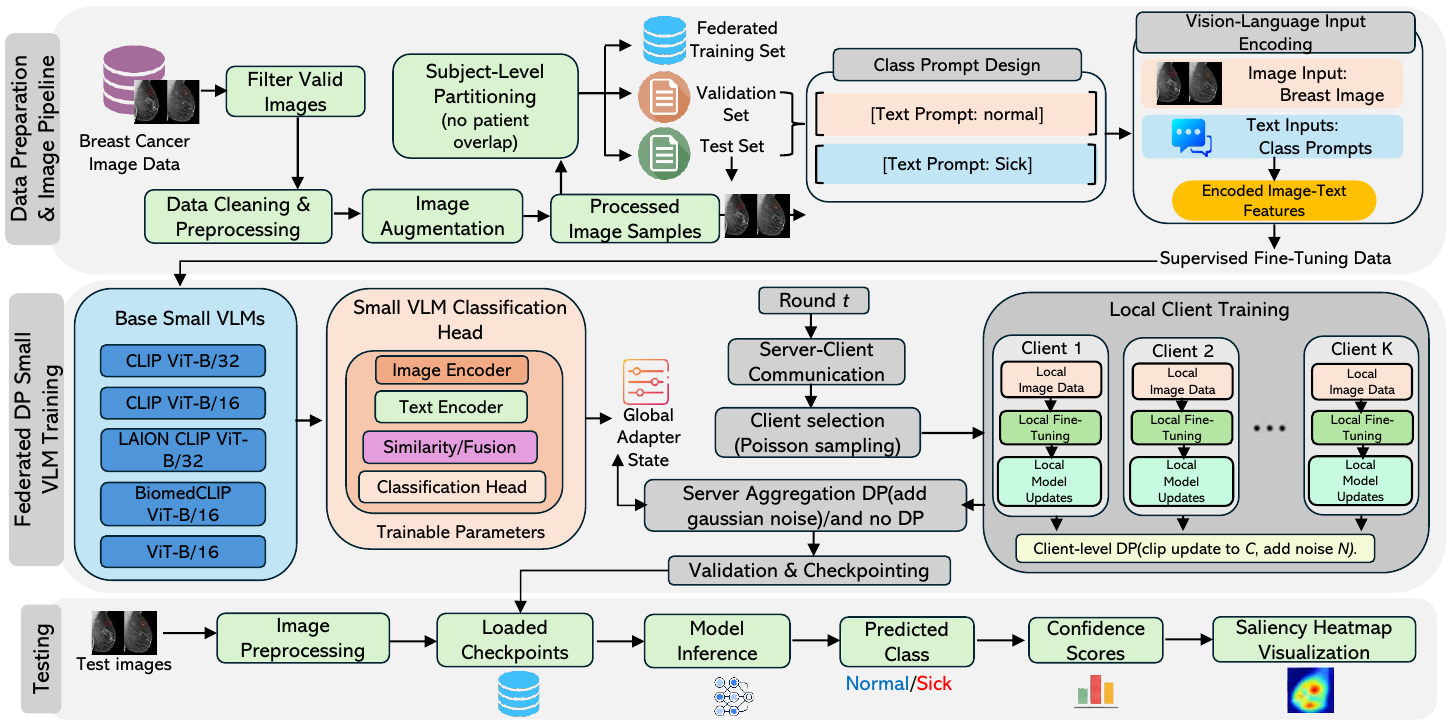}
\caption{Overview of the evaluated federated breast cancer classification framework, including data preparation, model training, optional client-level differential privacy, privacy-preserving aggregation, validation, and final evaluation.}
\label{fig:overall}
\end{figure*}

\begin{itemize}

\item We evaluate four compact CLIP-style vision-language models, including BiomedCLIP and an
ImageNet-pretrained ViT baseline for federated breast cancer classification across
two publicly available datasets, together with a controlled ablation that retains
the CLIP image encoder and discards only the text pathway, isolating the
contribution of language.

\item We integrate client-level differential privacy through update clipping
and Gaussian perturbation and analyze its privacy-utility trade-off using
formal privacy accounting and observed client-update statistics.

\item We conduct a matched comparison between vision-language and vision-only
representations under FL and FL+DP to determine whether language-guided
representations provide measurable benefits for supervised breast cancer
classification.

\item We perform patient-level evaluation using Receiver Operating Characteristic-Area Under the Curve (ROC-AUC) with bootstrap 95\%
confidence intervals and complement it with threshold-dependent metrics,
decision-threshold sensitivity, score-distribution analysis, and majority-class
baselines.

\item We analyze inter-model agreement, communication traffic, and training
time to characterize how privacy mechanisms and model architecture affect not
only predictive performance but also model behavior and federated-system cost.

\end{itemize}

The remainder of this paper is organized as follows. Section~\ref{sec:related} discusses the related work. Section~\ref{sec:method} presents the proposed method and experimental setup. Section~\ref{sec:results} reports the experimental results and evaluation. Section~\ref{sec:limitation} discusses the limitations and future research directions, and Section~\ref{sec:conclusion} concludes the paper.
\section{Related Works}\label{sec:related}
This study intersects breast imaging AI, vision-language models, federated learning, and differential privacy. While these areas have been studied individually, fewer works examine them together while also considering patient-level performance, privacy-utility trade-offs, and federated-system cost. This section reviews the most relevant prior work and identifies the research gap addressed by this study.

\subsection{Artificial Intelligence for Breast Imaging}
Artificial intelligence has increasingly been investigated for breast cancer screening and image-based decision support~\cite{chen2025ai}. Deep learning approaches~\cite{sabry2026ai} have been applied to mammography~\cite{amin2025systematic}, ultrasound, magnetic resonance imaging, and thermal imaging~\cite{parvin2026artificial} for tasks including classification~\cite{luo2024deep}, lesion detection, and segmentation~\cite{gao2026application}. Explainable AI has also attracted growing attention because clinical users require information about the image regions that influence model predictions~\cite{mashekova2025review}. Recent work has additionally investigated lightweight and distributed AI architectures for breast cancer applications. Split learning and federated learning have been explored as alternatives to centralized training, particularly when medical imaging data cannot be easily shared across organizations~\cite{wu2026application, raza2025federated}.

\subsection{Vision-Language Models (VLM) in Medical Imaging}

Vision-language models provide a mechanism for learning joint representations between medical images and textual information~\cite{wei2025mm}. Recent medical VLM research has investigated multimodal contrastive learning, prompt learning, few-shot adaptation, and medical-image segmentation~\cite{wei2025mm}. Mammo-CLIP adapted vision-language representations to mammography and demonstrated the potential of CLIP-style architectures for improving data efficiency in breast imaging~\cite{ghosh2024mammo}. Other work has investigated multi-granularity mammography representations and concept-level analysis of mammography-oriented CLIP models~\cite{du2025multi}. MedCLIPSeg extended CLIP-based representations to medical-image segmentation by introducing probabilistic vision-language adaptation~\cite{koleilat2026medclipseg}. BiomedCoOp investigated prompt learning for biomedical image classification and demonstrated that language-guided representations can improve learning when labeled medical images are limited~\cite{koleilat2025biomedcoop}. These studies demonstrate the potential of image-text representations in medical imaging; however, most work focuses on centralized training. Privacy-preserving distributed VLM training remains comparatively less explored~\cite{chen2025restyled}.

\subsection{Federated Learning and Differential Privacy in Medical Imaging}
Federated learning has been increasingly investigated for privacy-aware medical AI because it allows multiple clients or institutions to collaboratively optimize a model without directly sharing their raw patient data. Shukla et al.~\cite{shukla2025federated} combined federated learning and differential privacy for breast cancer diagnosis using the Breast Cancer Wisconsin Diagnostic dataset, demonstrating the feasibility of privacy-preserving collaborative learning for breast cancer classification. However, that study uses structured diagnostic features rather than breast thermography images; therefore, its reported performance is not directly comparable with the image-based classification tasks considered in the present study. Privacy-preserving medical-image federated learning has also been investigated together with uncertainty estimation, secure aggregation, and protection against model-update leakage~\cite{koutsoubis2025privacy}. Such studies demonstrate that retaining medical data locally does not by itself eliminate all privacy risks because gradients and model parameters may reveal information about local samples~\cite{alissa2026investigating}.

\subsection{Research Gap}

Previous studies have separately investigated artificial intelligence for
breast imaging~\cite{chen2026leveraging, zheng2025mv}, vision-language representations~\cite{zhai2026ordinal,cao2025mammovlm} in medical imaging,
federated learning~\cite{jimenez2023memory, dharani2024privacy, ahmed2026fscl}, and differential privacy. However, relatively limited work~\cite{shukla2025federated, makhanov2026federated}
has examined these components jointly in breast thermography while also
considering patient-level uncertainty, privacy-utility trade-offs,
architecture-level agreement, threshold sensitivity, and communication cost.

A further limitation of many small medical-imaging studies is that model
performance is often summarized using a single accuracy or F1-score, even
though such metrics may change substantially with the decision threshold and
may be unstable when the number of independent patients is small. In
addition, comparisons between VLM-based representations and simpler
vision-only baselines are not always included, making it difficult to
determine whether the language pathway provides measurable benefit for a
specific supervised classification task.

This study addresses these gaps by evaluating four compact CLIP-style
vision-language backbones and a vision-only ViT baseline under matched
federated and federated-with-DP settings. Performance is evaluated primarily
at the patient level using ROC-AUC and bootstrap confidence intervals, with
additional analysis of decision thresholds, inter-model agreement,
communication cost, and the effect of differential privacy on learned
rankings.

\section{Materials and Methods}\label{sec:method}
This section describes the experimental methodology used in this study. We first introduce the two breast imaging datasets and the subject-level partitions used for training, validation, and testing, followed by the image preprocessing and augmentation pipeline. We then describe the model backbones evaluated within the federated learning framework, including four vision-language models and two vision-only models. Next, we present the federated training setup, in which the preprocessed training data are distributed across 20 simulated clients under non-IID partitions and the selected backbones are trained collaboratively. We subsequently describe the client-level differential privacy mechanism used to extend the federated setting to FL+DP. Finally, we present the model-selection and evaluation protocol, including patient-level performance, uncertainty analysis, inter-model agreement, and communication cost.


\begin{figure*}[t]
\centering
\includegraphics[width=0.85\textwidth]{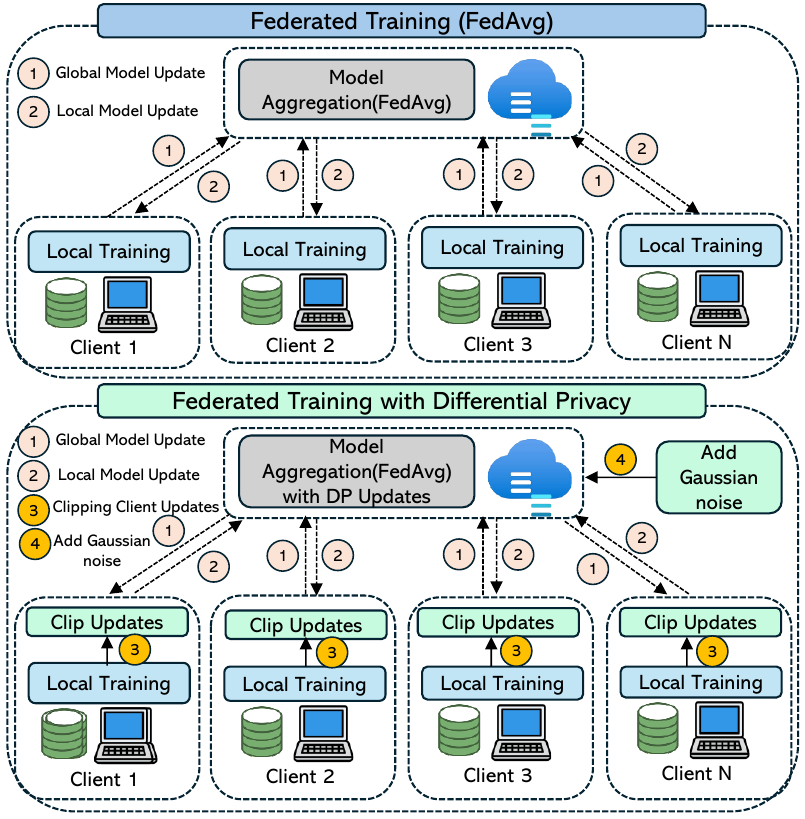}
\caption{Federated training workflows for standard FedAvg and differential privacy with update clipping and Gaussian noise.
}
\label{fig:fed_fed_diff}
\end{figure*}

\subsection{Datasets}
We utilized two publicly available breast imaging datasets to evaluate the proposed framework. Although the current experiments use thermographic images, the overall approach is not specific to this imaging modality and can be adapted to other breast imaging techniques, such as mammography, with appropriate preprocessing and task-specific data.

\underline{\textbf{\textit{BCD Dataset:}}} The first dataset, referred to as BCD in this study, is derived from the
DMR-IR breast thermography database. For the present experiments, cases with
unknown diagnostic status are excluded and the task is formulated as binary
classification between \textit{Normal} and \textit{Sick}. Subject-level
partitioning is used to prevent patient overlap across training, validation,
and test sets. The final held-out test partition contains 53 images from
52 patients~\cite{silva2014new}. 

\underline{\textbf{\textit{Breast Thermography Dataset:}}}
The second dataset contains infrared thermal images of the female thorax
for breast pathology analysis and includes data from 119 women, of whom
84 are benign and 35 are malignant. The task is binary classification
between \textit{Benign} and \textit{Malignant}. Subject-level partitioning
is applied to prevent patient overlap across training, validation, and test
sets. The held-out test partition contains 72 images from 24 patients,
with three views available for each patient. For patient-level evaluation,
the predicted probabilities from the three views are averaged before
computing ROC-AUC and the corresponding confidence intervals~\cite{rodriguez2024dataset}.

\subsection{Image Preprocessing}

We screen thermography images to exclude corrupted or unreadable files. The retained images are converted to RGB, resized to the input dimensions required by each CLIP architecture, and normalized according to the corresponding pretrained model's preprocessing specifications. Data augmentation and minority-class oversampling are restricted to the training partitions to increase sample diversity and mitigate class imbalance. The augmentation pipeline includes random cropping, horizontal flipping, rotation, color jitter, affine transformations, and random erasing. Minority-class oversampling is implemented by replicating the entries of minority-class training images in the training index rather than by storing duplicate image files. Because the augmentation pipeline is applied stochastically when each image is loaded, every replicated entry yields an independently augmented view in each epoch. The oversampled images are therefore not exact duplicates during training, and oversampling increases both the effective frequency and the augmented diversity of the minority class. Oversampling is performed after subject-level partitioning so that validation and test distributions remain unchanged, and the validation and test partitions receive only model-specific resizing and normalization.

\subsection{Backbones} Five pretrained backbones are evaluated under the same experimental protocol: OpenAI CLIP ViT-B/16~\cite{radford2021learning}, OpenAI CLIP ViT-B/32~\cite{radford2021learning}, LAION CLIP ViT-B/32~\cite{schuhmann2022laion}, BiomedCLIP ViT-B/16~\cite{zhang2023biomedclip}, a vision-only variant of OpenAI CLIP ViT-B/16, and an ImageNet-pretrained ViT-B/16 baseline~\cite{dosovitskiy2020image}. The first four models use pretrained image and text encoders to obtain visual and textual representations in a shared embedding space. For the BCD dataset, the class prompts correspond to ``normal'' and ``sick'', whereas the Breast Thermography dataset uses ``benign'' and ``malignant''. Each model additionally carries a lightweight classification head applied to the pooled image embedding, and the final logit combines the image-text similarity score with the head output. We use two controls to isolate different aspects of the vision–language design, and both are evaluated in terms of classification accuracy alone. vision-only CLIP ViT-B/16 variant retains the vision encoder and visual projection of OpenAI CLIP ViT-B/16 and discards the text encoder, replacing the image-text similarity term with a linear classification head. Because its visual representation is identical to that of the full CLIP ViT-B/16 model, the difference between the two configurations isolates the contribution of the text pathway. The ImageNet-pretrained ViT-B/16 shares the same architecture but derives its weights from supervised ImageNet classification rather than contrastive image-text pretraining, so comparing it with the vision-only CLIP variant isolates the effect of the pretraining corpus and objective. Trainable scope differs across the evaluated models. For the OpenAI and LAION CLIP backbones, the text encoder and text projection are frozen, while the vision encoder, the visual projection, the classification head and the two learned fusion weights are optimized. BiomedCLIP, the vision-only CLIP variant, and the ImageNet-pretrained ViT-B/16 are fine-tuned end-to-end together with their classification heads. The number of trainable parameters therefore varies across backbones, which should be considered when interpreting architectural comparisons. The consequences of this choice for the differentially private configuration, in which the communicated update spans the full model state rather than the trainable subset, are discussed in Section~\ref{sec:privacy}.

\begin{figure}[t]
    \centering
    \begin{subfigure}[t]{0.48\columnwidth}
        \centering\includegraphics[width=\linewidth]{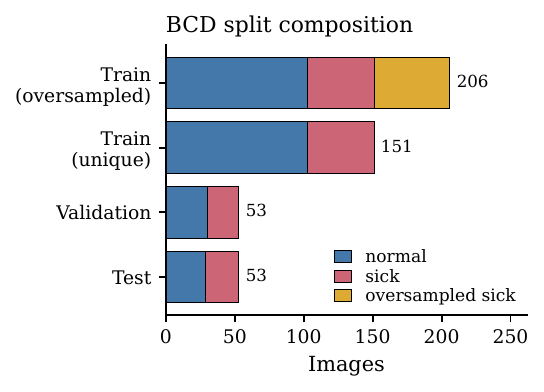}
        \caption{BCD: split composition.}\label{fig:bcd_splits}
    \end{subfigure}\hfill
    \begin{subfigure}[t]{0.48\columnwidth}
        \centering\includegraphics[width=\linewidth]{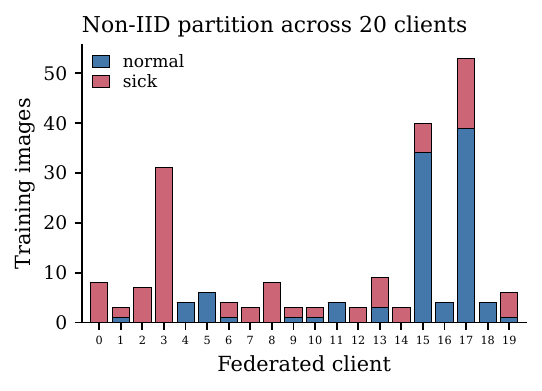}
        \caption{BCD: client partition.}\label{fig:bcd_clients}
    \end{subfigure}

    \vspace{0.6em}
    \begin{subfigure}[t]{0.48\columnwidth}
        \centering\includegraphics[width=\linewidth]{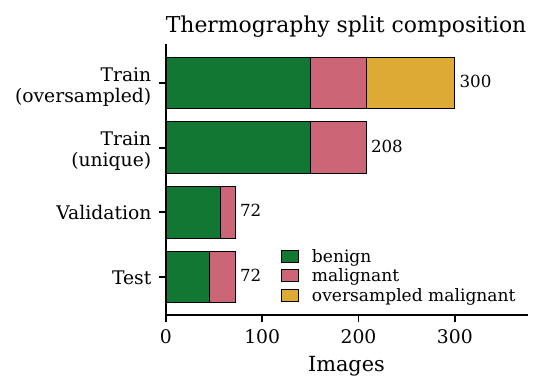}
        \caption{Thermography: split composition.}\label{fig:th_splits}
    \end{subfigure}\hfill
    \begin{subfigure}[t]{0.48\columnwidth}
        \centering\includegraphics[width=\linewidth]{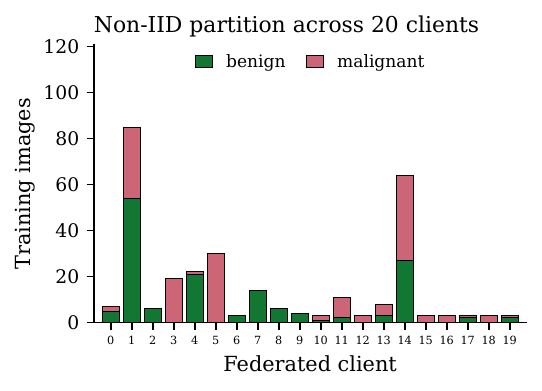}
        \caption{Thermography: client partition.}\label{fig:th_clients}
    \end{subfigure}
    \caption{Data partitioning. (a, c) Class composition of the subject-level
    splits, before and after minority-class oversampling. (b, d) Distribution of
    the training partition across 20 clients under Dirichlet allocation
    ($\alpha=0.5$ for BCD, $\alpha=1.0$ for breast thermography).}
    \label{fig:data_partition}
\end{figure}

\begin{figure}[t]
    \centering
    \begin{subfigure}[t]{0.48\columnwidth}
        \centering
        \includegraphics[width=\linewidth]{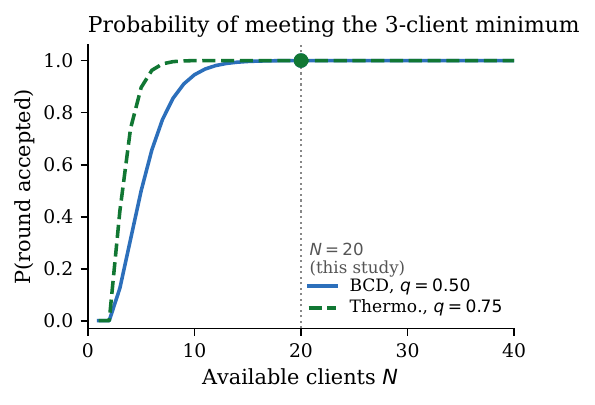}
        \caption{Round acceptance.}
        \label{fig:acceptance}
    \end{subfigure}
    \hfill
    \begin{subfigure}[t]{0.48\columnwidth}
        \centering
        \includegraphics[width=\linewidth]{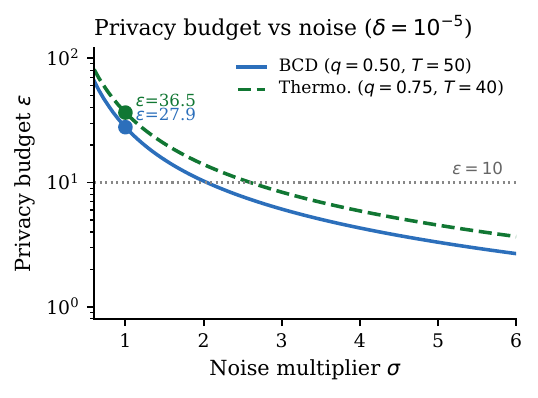}
        \caption{Privacy budget.}
        \label{fig:eps_sigma}
    \end{subfigure}
    \caption{Federated sampling and privacy accounting. (a) Probability of
    selecting at least three clients per round. (b) Accounted $\varepsilon$
    against noise multiplier $\sigma$ at $\delta=10^{-5}$.}
    \label{fig:sampling_privacy}
\end{figure}

\begin{algorithm}[t]
\caption{Federated Small-VLM Training with Optional DP}
\label{alg:fedvlm}

\textbf{Input:} Client datasets $\{D_k\}_{k=1}^{K}$, rounds $T$, sampling rate $q$,
minimum successful clients $m_{\min}$, clipping bound $C$, noise multiplier $\sigma$,
server learning rate $\eta_s$, privacy flag $\mathcal{P}$.

\textbf{Output:} Best global model $w^{*}$ and decision threshold $t^{*}$.

\begin{enumerate}
    \item Initialize global model $w_0$.

    \item For each round $t=0,\ldots,T-1$:
    \begin{enumerate}
        \item Sample clients $\mathcal{S}_t$ using Poisson sampling with rate $q$
        and broadcast $w_t$.

        \item For each client $k\in\mathcal{S}_t$:
        \begin{enumerate}
            \item Train locally and compute $\Delta w_{t,k}=w_{t,k}-w_t$;
            discard the update if it is not finite.

            \item If $\mathcal{P}=\mathrm{true}$, clip to the sensitivity bound
            \[
            \bar{\Delta}w_{t,k}
            =
            \Delta w_{t,k}
            \min\left(1,\frac{C}{\|\Delta w_{t,k}\|_2}\right);
            \]
            otherwise $\bar{\Delta}w_{t,k}=\Delta w_{t,k}$.
        \end{enumerate}

        \item Let $\mathcal{U}_t$ be the returned updates. If
        $|\mathcal{U}_t|<m_{\min}$, skip the round.

        \item Aggregate at the server. If $\mathcal{P}=\mathrm{true}$, perturb the
        \emph{sum} once and average uniformly:
        \[
        w_{t+1}
        =
        w_t+\frac{\eta_s}{|\mathcal{U}_t|}
        \left(
        \sum_{k\in\mathcal{U}_t}\bar{\Delta}w_{t,k}
        +\mathcal{N}(0,\sigma^2C^2I)
        \right);
        \]
        otherwise use weighted FedAvg,
        \[
        w_{t+1}
        =
        w_t+\eta_s
        \sum_{k\in\mathcal{U}_t}
        \frac{|D_k|}{\sum_{j\in\mathcal{U}_t}|D_j|}\,\bar{\Delta}w_{t,k}.
        \]

        \item Evaluate on the validation set, select the threshold maximizing
        macro F1, update the privacy accountant with $(\sigma,q)$, and store
        $(w^{*},t^{*})$ if the tuned macro F1 improves.
    \end{enumerate}

    \item Evaluate $w^{*}$ on the test set at both $0.5$ and $t^{*}$.
\end{enumerate}

\end{algorithm}

\subsection{Federated Training}

Federated training is conducted using simulated clients under the Federated Averaging (FedAvg) framework. The overall pipeline is summarized in Fig.~\ref{fig:overall}. The training partition of each dataset is distributed across \(N=20\) clients using Dirichlet allocation, with concentration parameter \(\alpha=0.5\) for the BCD dataset and \(\alpha=1.0\) for the breast thermography dataset. This produces a non-IID partition in which both the number of images and the class composition vary substantially across clients: client sizes range from 3 to 53 images for BCD and from 3 to 85 for breast thermography, and 12 and 11 clients respectively, contain images of a single class. The resulting partitions are shown in Fig.~\ref{fig:data_partition}.

Because minority-class oversampling is applied before partitioning, duplicated
copies of some minority-class images are allocated to different clients. Clients
are therefore not disjoint at the patient level: 27 of 151 patients in the BCD
training partition and 66 of 71 patients in the breast thermography training
partition contribute images to more than one client. This is a consequence of the
oversampling and partitioning order rather than of the federated protocol itself,
and its implications for the privacy guarantee are discussed in
Section~\ref{sec:privacy}.

Let the complete client set be \(\mathcal{K}=\{1,2,\ldots,N\}\). Rather than
requiring all clients to participate in every communication round, only a subset
of clients is selected. This partial-participation strategy is used for three main
reasons. First, it better reflects practical federated-learning environments, in
which all clients may not be simultaneously available because of differences in
connectivity, device availability, computational capacity, or local workload.
Second, selecting only a subset of clients reduces the communication and
computational cost of each round because the server does not need to exchange
model parameters with every client. Third, stochastic client participation
introduces variation in the participating client population across rounds,
allowing the global model to be updated from different subsets of distributed
local data over the course of training.

At each communication round \(t\), client participation is determined
independently using Poisson sampling. Each client \(k\in\mathcal{K}\) is selected
with probability \(q\), set to \(0.50\) for the BCD dataset and \(0.75\) for the
breast thermography dataset. The selection of a client \(k\) at round \(t\) can be
expressed using a Bernoulli random variable,

$$
I_{t,k}\sim \mathrm{Bernoulli}(q),
$$

where \(I_{t,k}=1\) indicates that client \(k\) is selected and \(I_{t,k}=0\)
indicates that it is not. The participating client set is therefore

$$
\mathcal{S}_t=\{k\in\mathcal{K}:I_{t,k}=1\},
$$

and the number of participating clients follows

$$
|\mathcal{S}_t|\sim \mathrm{Binomial}(N,q),
$$

with an expected number of selected clients given by

$$
\mathbb{E}[|\mathcal{S}_t|]=qN,
$$

corresponding to 10 clients per round for BCD and 15 for breast thermography.
The same sampling rate \(q\) is used by the privacy accountant described in
Section~\ref{sec:privacy}.

Because the sampling decision is made independently for every client and every
communication round, the exact number of selected clients varies from round to
round, and a client that participated in an earlier round may be selected again
later. When this occurs, the client performs additional local training on the same
locally assigned data; repeated participation across rounds is part of the
iterative optimization process. A communication round is accepted only when at
least three selected clients complete local training and return valid model
updates; otherwise that round is skipped and no global model is released.

At federated round \(t\), the server broadcasts the current global model parameters \(w_t\) to the participating clients in \(\mathcal{S}_t\). Each selected client \(k\) initializes its local model using \(w_t\). Local optimization uses AdamW with a learning rate of $5\times10^{-6}$ and weight decay of $0.01$, applied only to parameters with gradients enabled. Each selected client performs one local epoch per round with a batch size of 8 and shuffled sampling; a fresh optimizer is instantiated at the start of each client's local training, so optimizer state is not carried across rounds or shared between clients. Images are resized to $224\times224$ and normalized using the preprocessing statistics of the corresponding pretrained backbone. Training augmentation is applied only to the client partitions. The local objective is a focal loss with class weighting, computed as \[ \mathcal{L}^{\mathrm{task}}_k = \frac{1}{|B|}\sum_{i \in B}\, \alpha_{y_i}\,\bigl(1-p_{i,y_i}\bigr)^{\gamma}\, \bigl(-\log p_{i,y_i}\bigr), \] where $p_{i,y_i}$ is the predicted probability of the true class, $\alpha_{y_i}$ is the inverse-frequency class weight computed once from the pooled union of all client partitions and shared by every client rather than estimated locally, and $\gamma$ is the focusing parameter, set to $1.5$ for BCD and $1.0$ for breast thermography. Gradient clipping with a maximum $\ell_2$ norm of $1.0$ is applied to the trainable parameters at each optimization step; this is a stability measure during local training and is distinct from the client-update clipping used for differential privacy in Section~\ref{sec:privacy}. A FedProx proximal
term with coefficient \(\mu=0.001\) is added to the local objective,

$$
\mathcal{L}_k(w)=\mathcal{L}^{\mathrm{task}}_k(w)+\frac{\mu}{2}\lVert w-w_t\rVert_2^{2},
$$

which limits divergence of local solutions under the non-IID partition. The
resulting locally optimized parameters \(w_{t,k}\) give the client update

$$
\Delta w_{t,k}=w_{t,k}-w_t.
$$

The raw thermography images remain at the client side throughout training and are
not transmitted to the central server; only the resulting model updates are
communicated. In the non-private federated setting, the valid local updates are
combined using data-size-weighted FedAvg,

$$
\Delta w_t=
\sum_{k\in\mathcal{S}_t}
\frac{n_k}{\sum_{j\in\mathcal{S}_t}n_j}
\Delta w_{t,k},
$$

where \(n_k\) denotes the number of local training samples available at client
\(k\). The corresponding aggregation rule for the differentially private setting,
which uses uniform rather than size-weighted averaging, is given in
Section~\ref{sec:privacy}. The global model is subsequently updated as

$$
w_{t+1}=w_t+\eta_s\Delta w_t,
$$

where the server learning rate is set to \(\eta_s=1.0\). The number of
communication rounds is selected according to the observed convergence behavior of
each dataset, with 50 rounds used for the BCD dataset and 40 rounds for the breast
thermography dataset. The effective local learning rate is set to
\(5\times10^{-6}\), and all experiments use a fixed random seed of 42 to support
reproducibility.


\subsection{Client-Level Differential Privacy}\label{sec:privacy}

Federated learning prevents the direct centralization of raw images but does not eliminate the possibility of information leakage through communicated model updates. To provide an additional layer of privacy protection, we evaluate a client-level differential privacy mechanism based on update clipping and Gaussian perturbation. The private and non-private aggregation paths are contrasted in Fig.~\ref{fig:fed_fed_diff}. Client-level differential privacy bounds the influence of any single client's entire dataset on the released global model, rather than the influence of any individual training image.

\paragraph{Scope of the mechanism.} In the implementation evaluated here, the client update is formed from the complete model state rather than from the trainable subset. For a client $k$ at round $t$, the update $\Delta w_{t,k}=w_{t,k}-w_t$ is taken over every floating-point entry of the model state dictionary, and clipping, perturbation, and aggregation are all applied to that full vector. Consequently, parameters that are frozen during local optimization, the text encoder and text projection of the CLIP backbones, contribute exactly zero to $\Delta w_{t,k}$ but are nonetheless perturbed during aggregation and are written back into the global model. The frozen encoders therefore do \emph{not} remain unchanged under the private setting, although they do remain unchanged in the non-private setting, where a zero update averages to zero.

\paragraph{Mechanism.} Each client update is first clipped to a fixed $\ell_2$ norm bound, 
\[ \bar{\Delta}w_{t,k}=\Delta w_{t,k} \min\!\left(1,\frac{C}{\lVert\Delta w_{t,k}\rVert_2}\right), \] where $C$ is the clipping threshold. The server then sums the clipped updates, adds a single Gaussian perturbation, and averages uniformly over the participating clients, \[ w_{t+1}=w_t+\frac{\eta_s}{|\mathcal{S}_t|} \left(\sum_{k\in\mathcal{S}_t}\bar{\Delta}w_{t,k} +\mathcal{N}\!\left(0,\sigma^2C^2I\right)\right). \] A single perturbation of the aggregate, rather than one per client, matches the sensitivity assumed by the privacy accountant. Uniform averaging replaces the size-weighted scheme of the non-private setting because it bounds the contribution of any single client to the aggregate at $C$, which size-weighted averaging would not. Clipping is a client-side operation in principle and is applied at the aggregator in the present simulation; the two are equivalent because clipping is deterministic.

\paragraph{Privacy accounting.} The privacy parameters are $C=1.0$, $\sigma=1.0$ and $\delta=10^{-5}$. Accounting is performed under add/remove-client adjacency, giving $\varepsilon=27.9$ for the 50-round BCD experiments and $\varepsilon=36.5$ for the 40-round breast thermography experiments. All rounds were accepted in both settings: the three-client minimum was never approached (Fig.~\ref{fig:sampling_privacy}a), so every round contributed an aggregation and an accounting step. Fig.~\ref{fig:sampling_privacy}b shows how the accounted budget varies with the noise multiplier; $\varepsilon=10$ would require $\sigma\approx2.0$ for BCD and $\sigma\approx2.6$ for breast thermography. These values describe the accounted privacy loss of the evaluated configuration and should not be read as a strong clinical privacy guarantee.

\paragraph{Consequences of the chosen scope.} Applying the mechanism to the full model state has two measurable effects. First, the perturbation is element-wise with per-coordinate standard deviation $\sigma C$, so the noise vector has $\ell_2$ norm $\sigma C\sqrt{d}$, where $d$ is the total number of floating-point parameters. For the evaluated backbones $d\approx8.6\times10^{7}$ to $2.0\times10^{8}$, giving a noise norm of roughly $9\times10^{3}$ to $1.4\times10^{4}$, against a clipped update whose norm is bounded by $C=1.0$ and whose recorded value is $0.037$--$0.071$. The noise magnitude is therefore governed by the total model dimension rather than by the dimension of the trainable subset. Second, because frozen parameters receive this perturbation while contributing no signal, the pretrained text representations drift over training: after division by $|\mathcal{S}_t|$ the per-round perturbation of an individual weight has standard deviation $\sigma C/|\mathcal{S}_t|\approx0.10$ for BCD and $0.07$ for breast thermography, accumulating to approximately $0.70$ and $0.42$ over 50 and 40 rounds, which is an order of magnitude larger than typical pretrained weight magnitudes. 

These two effects, rather than the privacy guarantee itself, account for the performance decline reported in Section~\ref{sec:results}: under the private setting the models lose their agreement with their non-private counterparts (Table~\ref{tab:agreement}), which is consistent with a global model dominated by perturbation rather than by learned updates. Restricting the update vector to parameters with \texttt{requires\_grad=True} would reduce $d$ by several orders of magnitude and leave the frozen encoders untouched at the same $\varepsilon$, and calibrating $C$ to the observed update norms would reduce the injected noise further at no privacy cost. We report the present configuration as evaluated and identify these as the principal corrections required before the privacy--utility trade-off of this framework can be assessed; they are not addressed in the current experiments. The complete federated training procedure is summarized in Algorithm~\ref{alg:fedvlm}.

\subsection{Model Selection and Evaluation}

After each federated round, the global model is evaluated on the validation partition. The checkpoint obtaining the highest validation macro F1-score is retained for final evaluation. In addition, a decision threshold $t^{\ast}$ is selected using the validation data. Because multiple images may correspond to the same patient, the primary performance analysis is conducted at the patient level. For patients with multiple views, image-level predicted probabilities are averaged before computing patient-level metrics. Patient-level ROC-AUC and its confidence intervals are computed after aggregating predictions within each patient. Macro F1-score, sensitivity, and specificity are computed at the image level unless otherwise stated. Patient-level ROC-AUC is used as the primary metric because it evaluates ranking performance independently of a fixed classification threshold.

To examine whether different architectures assign similar rankings to the same test cases, we also compute pairwise Spearman correlations between the predicted probabilities produced by the five backbones under each training configuration. Communication traffic and training time are reported where available to provide an additional comparison of predictive utility and federated-system cost.

\paragraph{AI-assisted copy editing}AI tools were used solely to improve grammar, clarity, and readability of the manuscript; all scientific content, interpretations, and conclusions were reviewed and verified by the authors.
\section{Results and Discussion}\label{sec:results} Tables~\ref{tab:bcd} and~\ref{tab:thermo} summarize the predictive performance of the evaluated backbones under federated learning (FL) and federated learning with client-level differential privacy (FL+DP). We use patient-level ROC-AUC with bootstrap 95\% confidence intervals as the primary evaluation metric. Macro F1-score, sensitivity, and specificity are reported as complementary threshold-dependent measures. Additional analyses examine the effect of differential privacy (Table~\ref{tab:dpeffect}), agreement among independently pretrained backbones (Table~\ref{tab:agreement}), and communication and computation cost (Table~\ref{tab:cost}).

\subsection{Clinical Context and Scope of the Two Tasks} \label{sec:clinical_context} The two classification tasks are not clinically equivalent, and this should be kept in mind when comparing performance between them. In the BCD dataset, the positive class is labelled \emph{Sick}. The released annotations do not resolve this category into benign and malignant disease, so it may encompass malignant lesions together with benign pathology. The BCD task is therefore closer to abnormal-versus-normal discrimination than to cancer detection. The Breast Thermography dataset labels benign and malignant cases separately, so its task is the clinically harder distinction between two abnormal categories rather than between presence and absence of abnormality. Neither dataset releases the clinical variables that would allow a more detailed analysis. Lesion size, tumour stage, receptor status, histopathological subtype, time from imaging to diagnosis, and patient age are not available, so severity-stratified and subgroup analyses are not possible here. Performance figures should accordingly be read as aggregate discrimination over heterogeneous positive classes rather than as stage-specific detection rates. Finally, thermography measures surface temperature patterns produced by vascular and metabolic activity, not tissue structure or density. High classification performance on these datasets therefore does not establish clinical effectiveness for cancer detection or screening. The U.S. Food and Drug Administration has cleared thermography devices only as an \emph{adjunctive} tool used alongside a primary screening test such as mammography, and states that thermography is not effective as a standalone test for breast cancer screening or for detecting early-stage disease~\cite{fda2023thermography}. We therefore report classification performance as a measure of whether the training framework preserves a learnable signal under privacy constraints, not as evidence of diagnostic accuracy.

\subsection{Performance on the BCD Dataset} 
\begin{figure*}[t]
\centering

\begin{subfigure}[t]{0.48\textwidth}
    \centering
    \includegraphics[width=\linewidth]{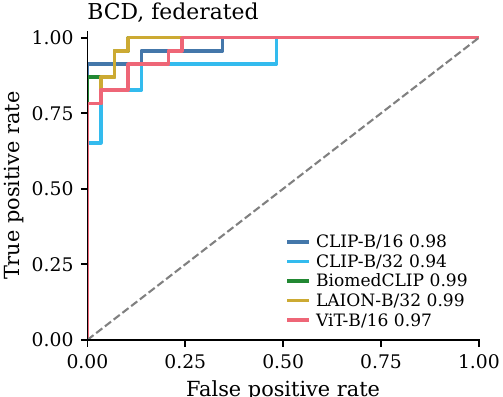}
    \caption{BCD, federated learning}
    \label{fig:roc_bcd_fl}
\end{subfigure}
\hfill
\begin{subfigure}[t]{0.48\textwidth}
    \centering
    \includegraphics[width=\linewidth]{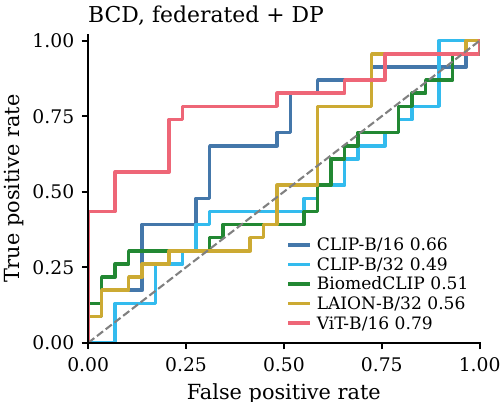}
    \caption{BCD, federated learning + DP}
    \label{fig:roc_bcd_dp}
\end{subfigure}

\vspace{0.8em}

\begin{subfigure}[t]{0.48\textwidth}
    \centering
    \includegraphics[width=\linewidth]{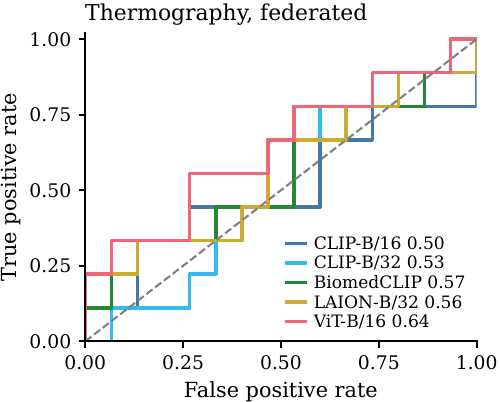}
    \caption{Breast thermography, federated learning}
    \label{fig:roc_thermo_fl}
\end{subfigure}
\hfill
\begin{subfigure}[t]{0.48\textwidth}
    \centering
    \includegraphics[width=\linewidth]{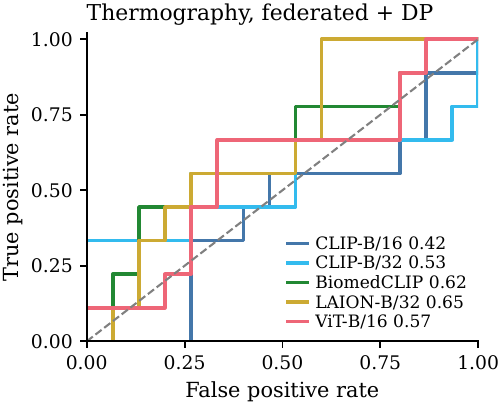}
    \caption{Breast thermography, federated learning + DP}
    \label{fig:roc_thermo_dp}
\end{subfigure}

\caption{
Patient-level ROC curves for the five evaluated backbones under federated
learning (FL) and federated learning with client-level differential privacy
(FL+DP). Panels (a) and (b) show the BCD normal-versus-sick task under FL and
FL+DP, respectively. Panels (c) and (d) show the corresponding results for the
Breast Thermography benign-versus-malignant task. Values reported in the
legends denote patient-level ROC-AUC.
}
\label{fig:roc_results}

\end{figure*}

The BCD normal-versus-sick task is consistently learned under federated training. As reported in Table~\ref{tab:bcd} and illustrated in Fig.~\ref{fig:roc_results}(a), all backbones achieve high patient-level ROC-AUC, ranging from 0.940 to 0.990. BiomedCLIP obtains the highest point estimate of 0.990, followed by LAION CLIP ViT-B/32 at 0.987, OpenAI CLIP ViT-B/16 at 0.979, the vision-only ViT-B/16 baseline at 0.970, and OpenAI CLIP ViT-B/32 at 0.940. All FL configurations yield $p<0.001$. The bootstrap confidence intervals in Fig.~\ref{fig:auc_privacy}(a) overlap considerably across architectures, so the numerical ordering does not provide strong evidence that any single backbone is superior. The score distributions in Fig.~\ref{fig:model_diagnostics}(c) show that normal and sick cases are separated along the predicted $p(\mathrm{sick})$ axis for all backbones, although the absolute score ranges differ substantially between architectures. This explains why ranking-based performance can remain high even when classification metrics at a fixed 0.5 threshold are poor. Section~\ref{sec:confound} examines what this learned signal is likely to be.

\subsection{Acquisition-Period Confounding in the BCD Dataset} \label{sec:confound} 
\begin{figure*}[t]

\centering

\includegraphics[width=\textwidth]{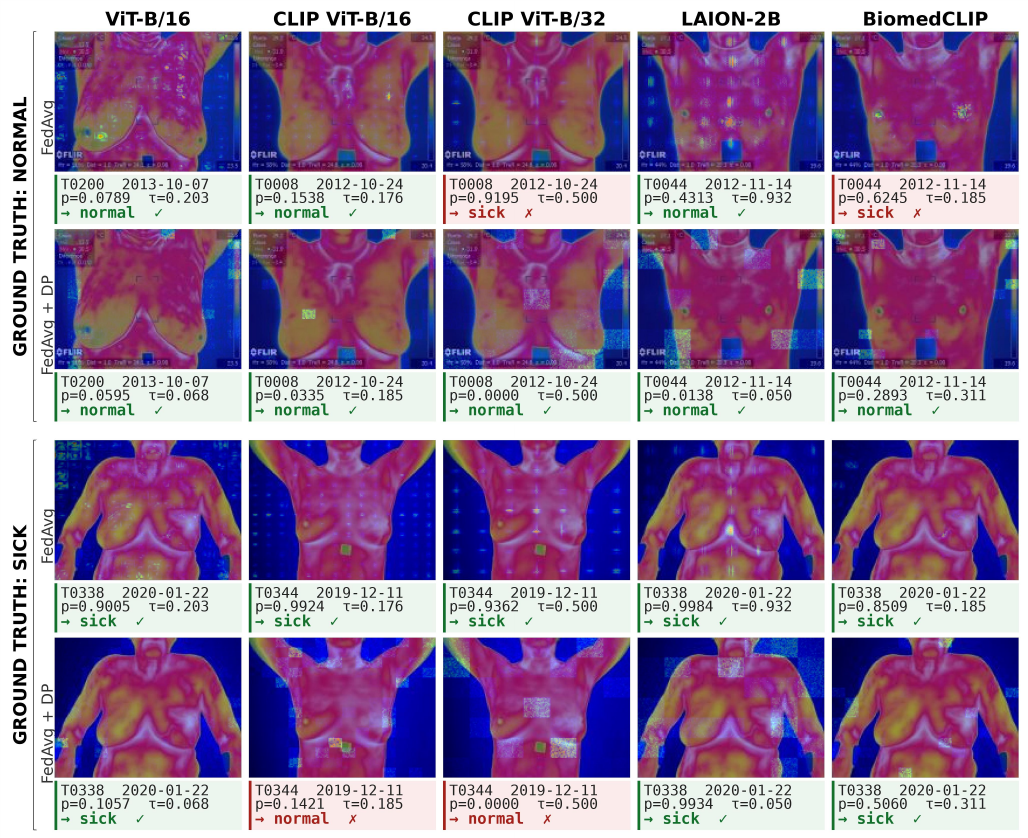}

\caption{Representative BCD predictions under FL and FL+DP across the five evaluated backbones.}

\label{fig:bcd_federated_predictions}

\end{figure*}

The BCD file names encode acquisition dates, which permits a direct test of whether the observed discrimination reflects acquisition conditions rather than thermal pathology. In the test partition, all images acquired in 2012 and 13 of 15 acquired in 2013 are normal, whereas every image acquired from 2015 onwards is sick; normal cases span 2012--2014 and sick cases 2013--2020, overlapping in only two years. Acquisition year alone attains a patient-level ROC-AUC of \textbf{0.974}, indistinguishable from the 0.934--0.990 obtained by the evaluated backbones, and model scores correlate with year at Spearman $\rho=0.76$--$0.83$ against $\rho=0.75$--$0.84$ with the diagnostic label. Restricting evaluation to the two overlapping years leaves only 14 normal and 7 sick images, within which year alone still attains 0.821, so the two effects cannot be separated with the available data. Consistent with this, the representative predictions in Fig.~\ref{fig:bcd_federated_predictions} show systematic differences in framing, field of view, and posture between the normal and sick examples. We therefore cannot exclude that the high BCD performance reflects an acquisition-period signature rather than thermal features of disease. The high inter-model agreement reported in Section~\ref{sec:agreement} is consistent with this interpretation, since a salient global cue is what architecturally distinct backbones would be expected to converge on. Addressing the confound would require acquisition-matched preprocessing or a cohort in which class and acquisition period are not confounded, neither of which is possible with the released data. We therefore report the BCD results as an evaluation of the training framework in the presence of a learnable signal, without asserting that the signal is physiological.

\subsection{Clinical Context and Scope of the Two Tasks} \label{sec:clinical_context} The two tasks are not clinically equivalent. In BCD the positive class is labelled \emph{Sick}, a category the released annotations do not resolve into benign and malignant disease, so the task is closer to abnormal-versus-normal discrimination than to cancer detection. The Breast Thermography dataset labels benign and malignant separately, requiring the harder distinction between two abnormal categories. Neither dataset releases lesion size, stage, receptor status, subtype, time to diagnosis, or patient age, so severity-stratified and subgroup analyses are not possible. Thermography measures surface temperature patterns produced by vascular and metabolic activity, not tissue structure. High classification performance on these datasets therefore does not establish clinical effectiveness for cancer detection or screening. The U.S. Food and Drug Administration has cleared thermography devices only as an \emph{adjunctive} tool used alongside a primary screening test such as mammography, and states that it is not effective as a standalone screening test~\cite{fda2023thermography}. We accordingly report classification performance as a measure of whether the training framework preserves a learnable signal under privacy constraints, not as evidence of diagnostic accuracy.

\subsection{Performance on the Breast Thermography Dataset} 
\begin{figure*}[t]
\centering

\begin{subfigure}[t]{0.48\textwidth}
    \centering
    \includegraphics[width=\linewidth]{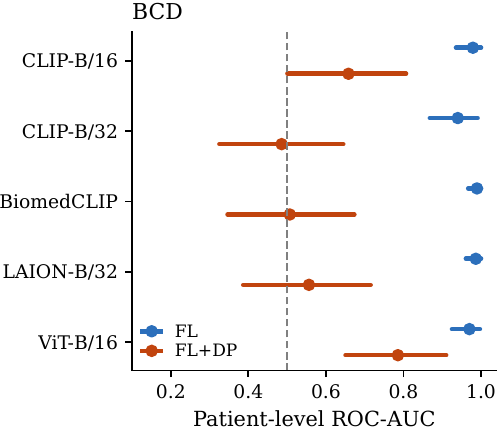}
    \caption{BCD}
    \label{fig:forest_bcd}
\end{subfigure}
\hfill
\begin{subfigure}[t]{0.48\textwidth}
    \centering
    \includegraphics[width=\linewidth]{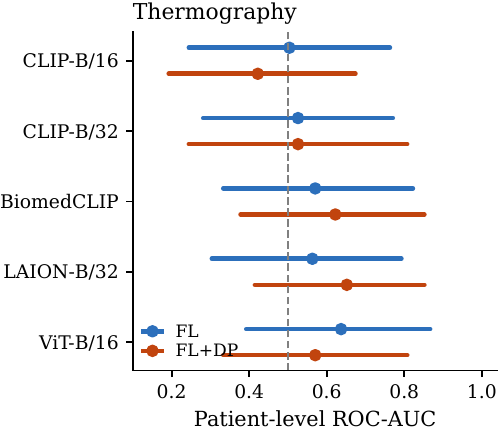}
    \caption{Breast Thermography}
    \label{fig:forest_thermo}
\end{subfigure}

\vspace{0.3em}

\begin{subfigure}[t]{0.48\textwidth}
    \centering
    \includegraphics[width=\linewidth]{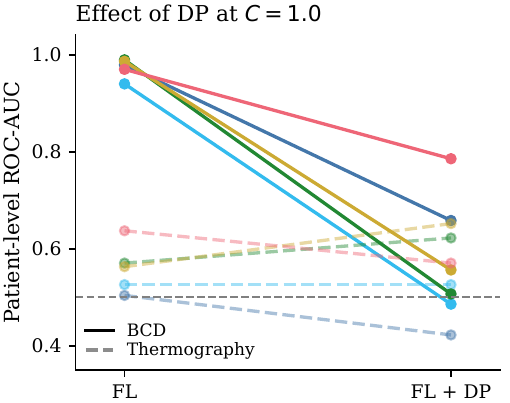}
    \caption{Effect of DP}
    \label{fig:dp_effect}
\end{subfigure}

\caption{Patient-level discrimination and the effect of client-level
differential privacy. Panels (a) and (b) show ROC-AUC estimates with
bootstrap 95\% confidence intervals for BCD and Breast Thermography,
respectively. The dashed vertical line denotes a chance-level AUC of 0.50.
Panel (c) shows the change in patient-level ROC-AUC after introducing
client-level DP with $C=1.0$, $\sigma=1.0$, and $\delta=10^{-5}$.
Solid lines represent BCD, and dashed lines represent breast thermography.}
\label{fig:auc_privacy}

\end{figure*}

The benign-versus-malignant task produces substantially more uncertain results. Under FL, patient-level ROC-AUC ranges from 0.504 to 0.637 (Table~\ref{tab:thermo}; Fig.~\ref{fig:roc_results}(c)). The vision-only ViT-B/16 baseline obtains the largest point estimate at 0.637, followed by BiomedCLIP at 0.570, LAION CLIP ViT-B/32 at 0.563, OpenAI CLIP ViT-B/32 at 0.526, and OpenAI CLIP ViT-B/16 at 0.504. Every bootstrap 95\% confidence interval contains the chance level of 0.50 (Fig.~\ref{fig:auc_privacy}(b)), so the numerical differences among backbones should not be interpreted as a ranking of model quality. The test cohort of 24 independent patients, comprising 9 malignant and 15 benign cases, provides limited statistical resolution.

\begin{figure*}[t]
\centering

\begin{subfigure}[t]{0.48\textwidth}
    \centering
    \includegraphics[width=\linewidth]{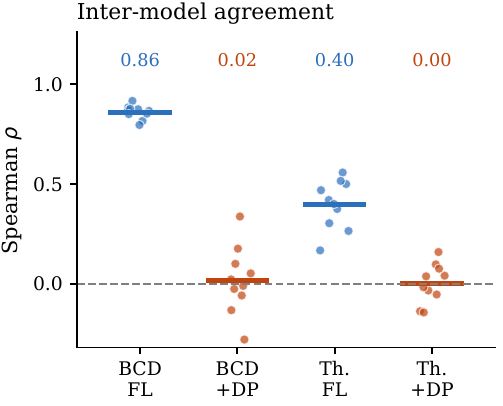}
    \caption{Inter-model agreement}
    \label{fig:agreement}
\end{subfigure}
\hfill
\begin{subfigure}[t]{0.48\textwidth}
    \centering
    \includegraphics[width=\linewidth]{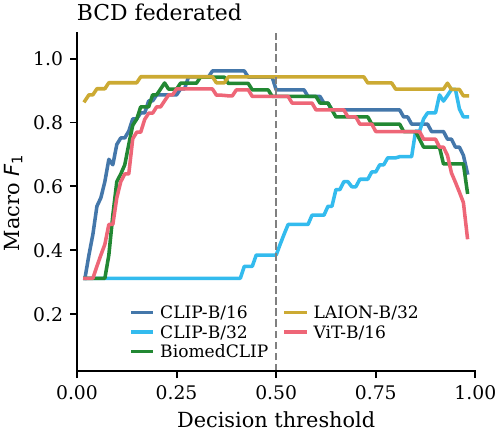}
    \caption{Threshold sensitivity on BCD}
    \label{fig:threshold}
\end{subfigure}

\vspace{0.5em}

\begin{subfigure}[t]{0.48\textwidth}
    \centering
    \includegraphics[width=\linewidth]{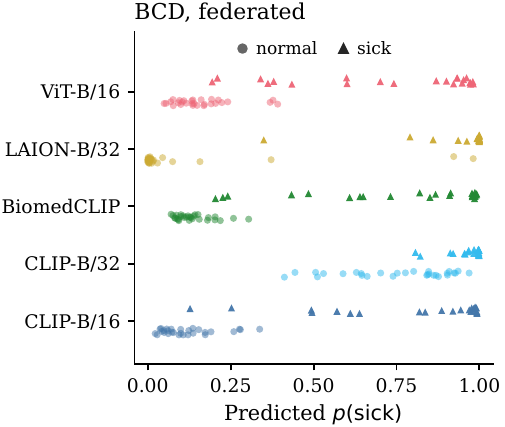}
    \caption{BCD score separation, federated learning}
    \label{fig:scores_fl}
\end{subfigure}
\hfill
\begin{subfigure}[t]{0.48\textwidth}
    \centering
    \includegraphics[width=\linewidth]{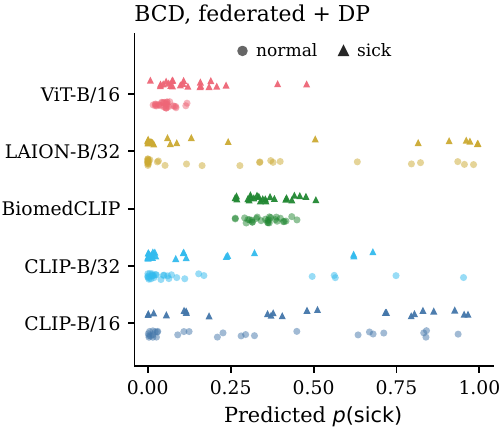}
    \caption{BCD score separation, federated learning + DP}
    \label{fig:scores_dp}
\end{subfigure}

\caption{
Diagnostic analysis of model behavior. Panel (a) reports pairwise
Spearman agreement between the predictions of the five evaluated
backbones. Panel (b) shows the sensitivity of image-level macro
F1-score to the decision threshold for the BCD federated models; the
dashed vertical line indicates the default threshold of 0.5. Panels
(c) and (d) show the distributions of predicted sick probabilities for
normal and sick BCD cases under federated learning (FL) and federated
learning with client-level differential privacy (FL+DP), respectively.
}
\label{fig:model_diagnostics}
\end{figure*}

\begin{figure}[t]
\centering
\includegraphics[width=\columnwidth]{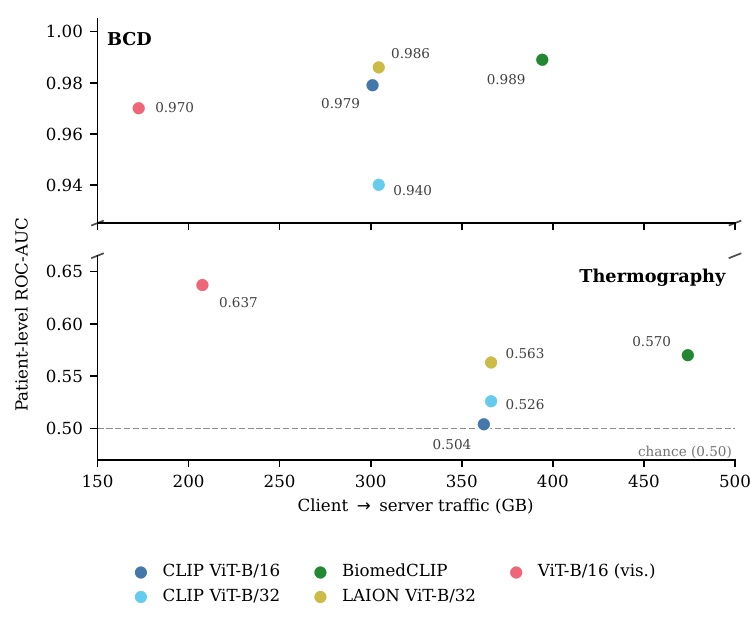}
\caption{Patient-level ROC-AUC versus client-to-server communication
traffic for federated configurations with available communication
measurements. Marker color identifies the model backbone, while marker
shape distinguishes the BCD and Breast Thermography datasets.}
\label{fig:cost_utility}
\end{figure}

\subsection{Comparison Between the Two Datasets} The substantial difference in performance between the two datasets is the most consistent observation of this study, and the analysis in Section~\ref{sec:confound} changes how it should be interpreted. The most likely explanation is not that thermal diagnosis is intrinsically easier on BCD. In BCD, the diagnostic class is almost perfectly aligned with the acquisition period, and a date stamp alone reproduces the performance of the trained models. The Breast Thermography dataset carries no comparable shortcut: benign and malignant cases were collected within a single study under a common protocol, so a model must rely on thermal image content. The contrast between the two datasets is therefore better understood as the presence of an acquisition-related shortcut in one dataset and its absence in the other, rather than as evidence that one diagnostic task is intrinsically easier than the other. Several secondary factors also differ and plausibly contribute. The two datasets define different classification problems, as described in Section~\ref{sec:clinical_context}: abnormal-versus-normal in BCD against benign-versus-malignant in breast thermography, the latter requiring discrimination between two abnormal categories and likely involving more subtle thermal differences. The thermography cohort provides only 24 independent test patients, which limits statistical resolution regardless of signal strength. Training-partition oversampling improves numerical class balance but cannot introduce the biological diversity that additional independent malignant patients would provide. Acquisition conditions, patient populations, and labeling criteria also differ between the two sources. None of these factors, individually or together, accounts for a gap of the observed magnitude as directly as the acquisition confound does.

\subsection{Effect of Client-Level Differential Privacy} The effect of client-level differential privacy differs substantially between the two datasets. Table~\ref{tab:dpeffect} and Fig.~\ref{fig:auc_privacy}(c) show a consistent reduction in BCD patient-level ROC-AUC after introducing DP, with a mean decrease of 0.374 AUC. The largest reduction occurs for BiomedCLIP, from 0.990 to 0.507, followed by OpenAI CLIP ViT-B/32 from 0.940 to 0.486, LAION CLIP ViT-B/32 from 0.987 to 0.556, and OpenAI CLIP ViT-B/16 from 0.979 to 0.658. The vision-only ViT-B/16 is the most robust configuration, decreasing from 0.970 to 0.786. The ROC curves in Fig.~\ref{fig:roc_results}(b) show several FL+DP models lying close to the chance diagonal, and the score distributions become considerably more overlapping (Fig.~\ref{fig:model_diagnostics}(d)). The magnitude of this degradation is consistent with the scope of the perturbation described in Section~\ref{sec:privacy}. Because the communicated update spans the full model state rather than the trainable subset, the injected noise has $\ell_2$ norm $\sigma C \sqrt{d}$ with $d\approx 10^{8}$, giving a per-coordinate signal-to-noise ratio of order $10^{-5}$, and frozen encoder parameters are perturbed although they contribute no signal. The observed loss should therefore be attributed to the evaluated implementation of the privacy mechanism rather than to client-level differential privacy in general. The breast thermography results must be interpreted differently. FL+DP AUC values range from 0.422 to 0.652, but all confidence intervals remain compatible with chance. The changes reported in Table~\ref{tab:dpeffect} range from $-0.081$ to $+0.089$, with a mean of approximately zero, and two of five backbones appear to improve. Because a privacy mechanism can only remove information, these changes provide a direct estimate of the measurement noise on this test set, roughly $\pm 0.08$ AUC, which exceeds every architectural difference observed. This illustrates a methodological point: the privacy-utility trade-off is interpretable only when the corresponding non-private model has first demonstrated reliable predictive performance.

\begin{figure*}[t]

\centering

\includegraphics[width=\textwidth]{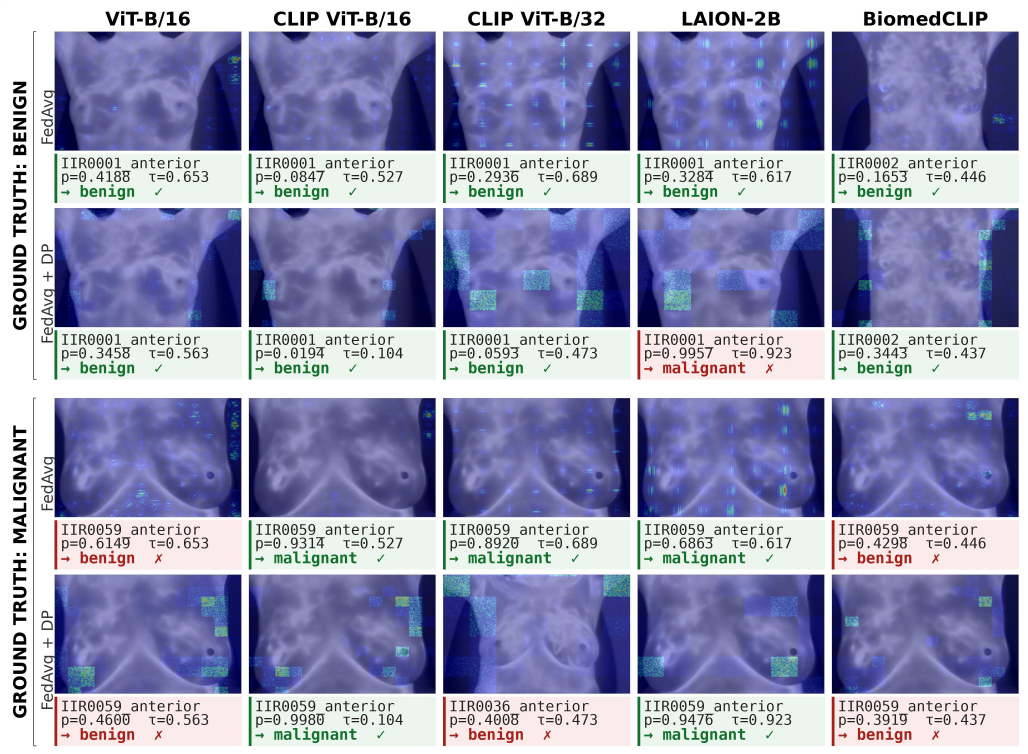}

\caption{Representative breast thermography predictions under FL and FL+DP across the five evaluated backbones.}

\label{fig:breast _federated_predictions}

\end{figure*}


\begin{table}[t]
\centering
\caption{Performance of the federated small VLMs on the BCD dataset (normal vs.\ sick;
53 test images, 52 patients; seed 42). AUC is patient-level with bootstrap 95\%
confidence intervals; $p$ from a one-sided Mann-Whitney test. Macro F1 is reported
at the default 0.5 boundary and at the validation-selected threshold $t^{\ast}$;
sensitivity and specificity are reported at the 0.5 boundary.}
\label{tab:bcd}
\small
\begin{tabular}{llccccccc}
\toprule
Setting & Model & AUC & 95\% CI & $p$ & F1$_{0.5}$ & F1$_{t^{\ast}}$ & Sens. & Spec. \\
\midrule
\multirow{5}{*}{\rotatebox{90}{FL}} & OpenAI CLIP ViT-B/16 & 0.979 & [0.94, 1.00] & $<$0.001 & 0.902 & 0.868 & 0.79 & 1.00 \\
 & OpenAI CLIP ViT-B/32 & 0.940 & [0.87, 0.99] & $<$0.001 & 0.385 & 0.867 & 1.00 & 0.07 \\
 & BiomedCLIP ViT-B/16 & 0.990 & [0.97, 1.00] & $<$0.001 & 0.882 & 0.887 & 0.75 & 1.00 \\
 & LAION CLIP ViT-B/32 & 0.987 & [0.96, 1.00] & $<$0.001 & 0.943 & 0.923 & 0.96 & 0.93 \\
 & ViT-B/16 (vision only) & 0.970 & [0.93, 1.00] & $<$0.001 & 0.882 & 0.849 & 0.75 & 1.00 \\
\midrule
\multirow{5}{*}{\rotatebox{90}{FL+DP}} & OpenAI CLIP ViT-B/16 & 0.658 & [0.50, 0.81] & 0.027 & 0.589 & 0.566 & 0.46 & 0.72 \\
 & OpenAI CLIP ViT-B/32 & 0.486 & [0.32, 0.65] & 0.573 & 0.430 & 0.450 & 0.12 & 0.86 \\
 & BiomedCLIP ViT-B/16 & 0.507 & [0.34, 0.67] & 0.471 & 0.398 & 0.408 & 0.04 & 1.00 \\
 & LAION CLIP ViT-B/32 & 0.556 & [0.39, 0.71] & 0.248 & 0.523 & 0.490 & 0.29 & 0.79 \\
 & ViT-B/16 (vision only) & 0.786 & [0.65, 0.91] & $<$0.001 & 0.354 & 0.752 & 0.00 & 1.00 \\
\midrule
\multicolumn{2}{l}{Majority-class baseline} & 0.500 & -- & -- & 0.354 & -- & 0.00 & 1.00 \\
\bottomrule
\end{tabular}
\end{table}


\begin{table}[t]
\centering
\caption{Performance on the Breast Thermography dataset
(benign vs.\ malignant; 72 test images, 24 patients, three views per patient;
seed 42). AUC is patient-level with bootstrap 95\% confidence intervals; $p$ from a
one-sided Mann-Whitney test. Macro F1 is reported at the default 0.5 boundary and
at the validation-selected threshold $t^{\ast}$; sensitivity and specificity are
reported at the 0.5 boundary. All confidence intervals include 0.50, indicating that
the available test cohort does not provide statistically conclusive evidence of
discrimination above chance for any evaluated configuration.}
\label{tab:thermo}
\small
\begin{tabular}{llccccccc}
\toprule
Setting & Model & AUC & 95\% CI & $p$ & F1$_{0.5}$ & F1$_{t^{\ast}}$ & Sens. & Spec. \\
\midrule
\multirow{5}{*}{\rotatebox{90}{FL}} & OpenAI CLIP ViT-B/16 & 0.504 & [0.24, 0.78] & 0.500 & 0.490 & 0.490 & 0.15 & 0.91 \\
 & OpenAI CLIP ViT-B/32 & 0.526 & [0.29, 0.78] & 0.429 & 0.485 & 0.485 & 0.59 & 0.42 \\
 & BiomedCLIP ViT-B/16 & 0.570 & [0.33, 0.81] & 0.296 & 0.486 & 0.486 & 0.11 & 0.98 \\
 & LAION CLIP ViT-B/32 & 0.563 & [0.30, 0.81] & 0.317 & 0.499 & 0.538 & 0.26 & 0.76 \\
 & ViT-B/16 (vision only) & 0.637 & [0.39, 0.87] & 0.142 & 0.637 & 0.519 & 0.59 & 0.69 \\
\midrule
\multirow{5}{*}{\rotatebox{90}{FL+DP}} & OpenAI CLIP ViT-B/16 & 0.422 & [0.19, 0.67] & 0.744 & 0.428 & 0.444 & 0.11 & 0.82 \\
 & OpenAI CLIP ViT-B/32 & 0.526 & [0.25, 0.81] & 0.429 & 0.545 & 0.545 & 0.22 & 0.91 \\
 & BiomedCLIP ViT-B/16 & 0.622 & [0.37, 0.85] & 0.170 & 0.385 & 0.515 & 0.00 & 1.00 \\
 & LAION CLIP ViT-B/32 & 0.652 & [0.41, 0.86] & 0.117 & 0.537 & 0.542 & 0.59 & 0.51 \\
 & ViT-B/16 (vision only) & 0.570 & [0.33, 0.81] & 0.296 & 0.596 & 0.551 & 0.56 & 0.64 \\
\midrule
\multicolumn{2}{l}{Majority-class baseline} & 0.500 & -- & -- & 0.385 & -- & 0.00 & 1.00 \\
\bottomrule
\end{tabular}
\end{table}


\begin{table}[t]
\centering
\caption{
Change in patient-level ROC-AUC after adding client-level differential
privacy ($C=1.0$, $\sigma=1.0$, $\delta=10^{-5}$).
$\rho_{\mathrm{FL,DP}}$ denotes the Spearman correlation between the
predictions of the corresponding FL and FL+DP models, and the mean is taken over
the five backbones. On BCD, the evaluated DP configuration substantially reduces
ranking performance. On breast thermography, the observed AUC changes are small
relative to the uncertainty of the patient-level evaluation.
}
\label{tab:dpeffect}
\small
\begin{tabular}{llcccc}
\toprule
Dataset & Model & AUC$_{\mathrm{FL}}$ & AUC$_{\mathrm{FL+DP}}$ & $\Delta$ & $\rho_{\mathrm{FL,DP}}$ \\
\midrule
\multirow{6}{*}{BCD} & OpenAI CLIP ViT-B/16 & 0.979 & 0.658 & $-0.321$ & $+0.403$ \\
 & OpenAI CLIP ViT-B/32 & 0.940 & 0.486 & $-0.454$ & $-0.004$ \\
 & BiomedCLIP ViT-B/16 & 0.990 & 0.507 & $-0.483$ & $+0.091$ \\
 & LAION CLIP ViT-B/32 & 0.987 & 0.556 & $-0.430$ & $-0.019$ \\
 & ViT-B/16 (ImageNet) & 0.970 & 0.786 & $-0.184$ & $+0.486$ \\
\cmidrule(l){2-6}
 & \textit{Mean} & & & $\mathbf{-0.401}$ & \\
\midrule
\multirow{6}{*}{Thermography} & OpenAI CLIP ViT-B/16 & 0.504 & 0.422 & $-0.081$ & $+0.155$ \\
 & OpenAI CLIP ViT-B/32 & 0.526 & 0.526 & $+0.000$ & $-0.058$ \\
 & BiomedCLIP ViT-B/16 & 0.570 & 0.622 & $+0.052$ & $+0.218$ \\
 & LAION CLIP ViT-B/32 & 0.563 & 0.652 & $+0.089$ & $-0.035$ \\
 & ViT-B/16 (ImageNet) & 0.637 & 0.570 & $-0.067$ & $+0.013$ \\
\cmidrule(l){2-6}
 & \textit{Mean} & & & $\mathbf{+0.002}$ & \\
\bottomrule
\end{tabular}
\end{table}


\begin{table}[t]
\centering
\caption{Pairwise Spearman correlation between the predictions of the five
architectures (10 pairs per group). Under federated training the models converge
on a common ranking; under DP at $C=1.0$ that agreement disappears on both
datasets.}
\label{tab:agreement}
\small
\begin{tabular}{llcccc}
\toprule
Dataset & Setting & Mean $\rho$ & Min & Max & Pairs $\rho>0$ \\
\midrule
\multirow{2}{*}{BCD} & Federated & 0.860 & $+0.80$ & $+0.92$ & 10/10 \\
 & Federated + DP & 0.018 & $-0.28$ & $+0.34$ & 5/10 \\
\midrule
\multirow{2}{*}{Thermography} & Federated & 0.397 & $+0.17$ & $+0.56$ & 10/10 \\
 & Federated + DP & 0.002 & $-0.14$ & $+0.16$ & 5/10 \\
\bottomrule
\end{tabular}
\end{table}


\begin{table}[t]
\centering
\caption{Communication and computation cost of the federated configurations, with
patient-level ROC-AUC reported for reference.}
\label{tab:cost}
\small
\begin{tabular}{lcccccc}
\toprule
 & \multicolumn{3}{c}{BCD} & \multicolumn{3}{c}{Thermography} \\
\cmidrule(lr){2-4}\cmidrule(lr){5-7}
Model & AUC & Traffic (GB) & Time (s) & AUC & Traffic (GB) & Time (s) \\
\midrule
OpenAI CLIP ViT-B/16 & 0.979 & 301 & 389 & 0.504 & 362 & 510 \\
OpenAI CLIP ViT-B/32 & 0.940 & 304 & 371 & 0.526 & 366 & 519 \\
BiomedCLIP ViT-B/16 & 0.990 & 394 & 484 & 0.570 & 474 & 633 \\
LAION CLIP ViT-B/32 & 0.987 & 304 & 376 & 0.563 & 366 & 474 \\
ViT-B/16 (vision only) & 0.970 & 173 & 290 & 0.637 & 208 & 379 \\
\bottomrule
\end{tabular}
\end{table}

\begin{table}[t]
\centering
\caption{Effect of removing the text encoder from CLIP ViT-B/16. Both
configurations use the same pretrained image encoder and visual projection; the
vision-only variant replaces the image--text similarity term with a linear
classification head. AUC is patient-level with bootstrap 95\% confidence intervals.}
\label{tab:textablation}
\small
\begin{tabular}{llccc}
\toprule
Dataset & Configuration & AUC & 95\% CI & Traffic (GB) \\
\midrule
\multirow{2}{*}{BCD} & CLIP ViT-B/16 & 0.979 & [0.94, 1.00] & 301 \\
 & \quad without text encoder & 0.993 & [0.97, 1.00] & 173 \\
\midrule
\multirow{2}{*}{Thermography} & CLIP ViT-B/16 & 0.504 & [0.24, 0.78] & 362 \\
 & \quad without text encoder & 0.585 & [0.33, 0.83] & 209 \\
\bottomrule
\end{tabular}
\end{table}

\subsection{Inter-Model Agreement} \label{sec:agreement} Inter-model agreement measures how similarly different backbones rank the same test samples. Table~\ref{tab:agreement} and Fig.~\ref{fig:model_diagnostics}(a) report pairwise Spearman correlations. Under federated training on BCD, the mean pairwise correlation is 0.860 with all 10 pairs positive. On breast thermography it is 0.397, again with all 10 pairs positive, which is notable because no individual model achieves a confidence interval excluding 0.50: the backbones produce a broadly similar ranking even where individual discrimination is not statistically established. The high BCD agreement should be read together with Section~\ref{sec:confound}. Convergence of architecturally distinct backbones is expected when a salient, easily learned cue is present, and an acquisition-period signature is such a cue. Agreement therefore indicates the presence of a shared learnable signal, not its clinical validity. After DP is introduced, agreement collapses on both datasets, from 0.860 to 0.018 on BCD and from 0.397 to 0.002 on breast thermography, with only 5 of 10 pairwise correlations remaining positive in each case. ROC-AUC measures consistency with the true labels, whereas inter-model correlation measures whether different architectures preserve a common ranking; the collapse of the latter on both datasets, including the one where the effect on AUC is not visible, makes it a more sensitive indicator of a perturbation-dominated model than accuracy alone.

\subsection{Decision-Threshold Sensitivity} Threshold-dependent metrics can vary substantially even when the underlying ranking ability of a model is unchanged (Fig.~\ref{fig:model_diagnostics}(b)). OpenAI CLIP ViT-B/32 on BCD achieves a patient-level AUC of 0.940, but its macro F1-score at the default 0.5 threshold is 0.385, rising to 0.867 at the validation-selected threshold $t^{\ast}$. The ImageNet-pretrained ViT-B/16 under FL+DP retains an AUC of 0.786 while its macro F1 at the default threshold is 0.354, because the model produces no positive predictions at that boundary; at $t^{\ast}$ it rises to 0.752. A low F1-score at a fixed threshold therefore does not imply a lack of ranking ability and may instead reflect score calibration relative to the chosen boundary. For this reason, we treat ROC-AUC as the primary metric, and we always interpret macro F1, sensitivity, and specificity together with the threshold used to obtain them and the corresponding confusion matrix.

\subsection{Vision-Language Models Versus Vision-Only Baselines} \label{sec:vlm_vs_vision} To determine whether the text encoder in CLIP contributes to breast image classification, we evaluate a vision-only variant of OpenAI CLIP ViT-B/16. This variant retains the pretrained vision encoder and visual projection of the full model but removes the text encoder and replaces the image-text similarity term with a linear classification head. Because the visual representation is unchanged, the comparison separates the contribution of the text pathway from that of the pretrained visual representation. We further compare both configurations with the vision-only ViT-B/16 baseline, which uses ImageNet-supervised pretraining, to examine whether CLIP-style image-text pretraining offers an advantage over conventional visual pretraining. The results are summarized in Table~\ref{tab:textablation}. On BCD, the vision-only CLIP variant achieves a patient-level ROC-AUC of 0.993 (95\% CI: 0.97--1.00), compared with 0.979 (0.94--1.00) for the full CLIP model and 0.970 (0.93--1.00) for the ImageNet-pretrained baseline. Although the vision-only CLIP variant yields the highest point estimate, the confidence intervals overlap substantially, so the differences are not statistically supported. The same pattern holds on the Breast Thermography dataset, where the three configurations achieve AUCs of 0.585, 0.504, and 0.637 respectively, all of which remain compatible with chance-level discrimination. The communication measurements clarify the trade-off. Removing the text encoder reduces client-to-server traffic on BCD from 301\,GB to 173\,GB, indicating that the language pathway imposes a substantial communication cost without a corresponding predictive benefit in the supervised binary classification tasks considered here. These results suggest that the pretrained visual representation accounts for most of the useful signal in this setting. The language pathway may nonetheless be valuable for tasks not evaluated in this study, such as zero-shot classification, prompt-based inference, image-report alignment, and open-vocabulary prediction. The strong BCD performance should also be interpreted cautiously in light of the acquisition-related confounding discussed in Section~\ref{sec:confound}, since higher discrimination does not necessarily imply stronger diagnostic capability.

\subsection{Communication and Computational Cost} Table~\ref{tab:cost} and Fig.~\ref{fig:cost_utility} compare predictive utility with federated communication requirements. On BCD, the vision-only ViT-B/16 achieves an AUC of 0.970 while requiring approximately 173\,GB of client-to-server traffic, against 301\,GB for OpenAI CLIP ViT-B/16, 304\,GB for OpenAI CLIP ViT-B/32 and LAION CLIP ViT-B/32, and 394\,GB for BiomedCLIP. It also requires less training time, completing in 290\,s against 371--484\,s for the CLIP-style models. The ablation reported in Section~\ref{sec:vlm_vs_vision} shows that this reduction is attributable to the text tower rather than to the visual architecture: removing the text encoder from CLIP ViT-B/16 while retaining its vision encoder reduces traffic to the same 173\,GB. This difference follows directly from the update scope described in Section~\ref{sec:privacy}. Because the communicated update spans the full model state, the frozen text encoder of the CLIP backbones is transmitted every round as a vector of zeros. For CLIP ViT-B/16 the text tower accounts for approximately 40\% of the parameters, so a substantial fraction of the reported traffic carries no information. Restricting communication to parameters with gradients enabled would remove most of the gap between the vision-only and vision--language configurations. These system-level differences should not be interpreted as evidence of superior clinical utility, particularly on breast thermography, where no configuration provides statistically conclusive discrimination.

\subsection{Overall Interpretation}

Taken together, the results demonstrate that no single metric is sufficient
for evaluating privacy-preserving federated breast cancer
classification. The ROC curves in Fig.~\ref{fig:roc_results} establish the
difference between a clearly learnable BCD task and a statistically uncertain
Breast Thermography task. The patient-level confidence intervals and DP-effect
analysis in Fig.~\ref{fig:auc_privacy} quantify uncertainty and show that the
evaluated privacy configuration substantially degrades the learned BCD
ranking. The model-behavior analyses in Fig.~\ref{fig:model_diagnostics}
show that this degradation is accompanied by reduced class-score separation
and a collapse in agreement among independently pretrained backbones, while
also demonstrating that threshold-dependent metrics can differ substantially
from ranking-based performance. Finally, Fig.~\ref{fig:cost_utility} shows
that the vision-only baseline achieves a favorable utility-to-communication
trade-off.

Three methodological conclusions follow from these observations. First,
patient-level evaluation is essential when multiple images originate from the
same individual, because treating correlated views as independent test samples
can overstate statistical evidence. Second, confidence intervals and explicit
chance or majority-class baselines are necessary when evaluating small
medical-imaging datasets; point estimates alone can create apparent
architectural differences that are not statistically supported. Third,
differential privacy should be evaluated together with update-scale
measurements and communication behavior rather than treated only as an
additional training switch. The privacy parameters, update norm, and model
architecture jointly determine the practical privacy-utility trade-off. The findings should nevertheless be interpreted within the limitations of the
study. All experiments use a single seed, the federated clients are simulated,
the test cohorts are relatively small, and only one DP configuration is
evaluated. In particular, the broad confidence intervals on Breast
Thermography prevent strong conclusions about architecture ranking or clinical
discrimination. The present results therefore establish methodological
behavior of the federated framework rather than clinical effectiveness of
thermography-based breast cancer screening.

\section{Limitations and Future Work}\label{sec:limitation}

The present study has several limitations that should be considered when interpreting the results. First, the number of independent patients in the held-out test sets is relatively small, with 52 patients in the BCD test split and only 24 patients in the Breast Thermography test split; this limitation is particularly important for the benign-versus-malignant task, where all bootstrap 95\% confidence intervals for patient-level ROC-AUC include 0.50, preventing strong conclusions about architectural superiority. Second, all experiments are conducted using a single random seed, so the reported bootstrap confidence intervals capture uncertainty associated with the held-out patient sample but not variability introduced by initialization, client sampling, augmentation, data ordering, or optimization. Third, the federated environment is simulated and does not fully capture practical multi-institutional conditions such as non-IID patient populations, unequal client sizes, heterogeneous imaging devices and acquisition protocols, intermittent participation, client dropout, network variability, and differences in computational resources. Fourth, the differential privacy experiments evaluate only one operating point with $C=1.0$, $\sigma=1.0$, and $\delta=10^{-5}$; because the observed client-update norms are substantially smaller than the clipping threshold, the reported FL+DP results may reflect a suboptimal privacy-utility configuration rather than the best achievable trade-off. In addition, although formal client-level privacy accounting is included, the resulting privacy budgets are relatively large and should therefore be interpreted as accounted privacy rather than a strong deployment-level guarantee. The study also does not include a centralized baseline under matched data splits and optimization settings, making it difficult to isolate the performance cost introduced by federation itself, and it does not evaluate empirical privacy attacks such as gradient inversion, update reconstruction, or membership inference. Communication and runtime measurements are also incomplete for some Breast Thermography configurations. Furthermore, the vision--language evaluation relies on simple class prompts within a supervised binary classification setting, so the competitive performance of the vision-only baseline does not imply that language representations are generally unnecessary; richer clinical prompts, zero-shot and few-shot transfer, open-vocabulary classification, and image--report alignment may reveal benefits not examined here. Finally, gradient-based saliency maps are used only as qualitative inspection tools and should not be interpreted as clinically validated localization of pathological regions. Future work should therefore evaluate multiple random seeds, larger and multi-center patient cohorts, centralized and matched federated baselines, non-IID and heterogeneous client settings, multiple clipping thresholds and noise multipliers, stronger privacy-utility operating points, empirical privacy attacks, complete communication and computation profiling, richer vision-language tasks, and clinically validated explainability methods. Prospective evaluation with standardized acquisition protocols, confirmed reference diagnoses, calibration and subgroup analyses, and assessment by breast-imaging specialists will also be necessary before the framework can be considered for clinical screening-support applications.
\section{Conclusion}\label{sec:conclusion} This study presents a systematic evaluation of federated small vision-language models for privacy-preserving breast thermography classification, using five pretrained backbones and reporting patient-level ROC-AUC with confidence intervals as the primary metric. The two datasets behave very differently. The BCD normal-versus-sick task yields high patient-level discrimination under federated learning, but analysis of the released metadata shows that the diagnostic class is almost perfectly aligned with acquisition period: acquisition year alone attains a patient-level AUC of 0.974, indistinguishable from the trained models. The high BCD performance therefore cannot be attributed to thermal features of disease. The breast thermography benign-versus-malignant task, which carries no comparable shortcut, remains statistically uncertain, with every confidence interval including chance. An additional experiment removed the text encoder from CLIP ViT-B/16 while retaining its image encoder. The resulting vision-only variant matches the full model while halving client-to-server traffic, and an ImageNet-pretrained ViT of the same architecture performs comparably. For the supervised binary tasks considered here, neither the language pathway nor contrastive image-text pretraining confers a measurable predictive advantage. Client-level differential privacy substantially reduces BCD performance by 0.374 AUC on average, and collapses inter-model agreement on both datasets. This loss is attributable to the scope of the perturbation rather than to the privacy guarantee itself: the communicated update spans the full model state rather than the trainable subset, so the injected noise scales with total model dimension and frozen encoder parameters are perturbed despite contributing no signal. The clipping threshold is also far larger than the observed update norms. Restricting the update to trainable parameters and calibrating the threshold to the measured scale are the principal corrections required before the privacy-utility trade-off of this framework can be assessed. Overall, the findings demonstrate the technical feasibility of the evaluated federated training pipeline while identifying substantial limitations: small patient cohorts, a dataset shortcut that undermines the apparent diagnostic signal, sensitivity to the decision threshold, and communication overhead. The results characterise the methodological behaviour of the framework rather than the clinical effectiveness of thermography-based screening, for which thermography remains cleared only as an adjunctive tool~\cite{fda2023thermography}.

\section{Statements and Declarations}

\paragraph{Author Contribution}M.M.W.K.: conceptualization, methodology, investigation, experimental analysis, visualization, validation, writing the original draft, review and editing. N.Y.: methodology, investigation, supervision, review and editing. S.B.D.: investigation, clinical interpretation, validation, review and editing. M.Z.I.: investigation, clinical interpretation, validation, review and editing. A.R.S.: conceptualization, methodology, supervision, project administration, validation, review and editing. All authors read and approved the final manuscript.

\paragraph{Funding:}
The authors received no specific funding for this work.

\paragraph{Data Availability}
The datasets analyzed in this study are publicly available from the
respective sources cited in the manuscript.

\paragraph{Ethics Approval}This study used publicly available data.
Therefore, no ethical approval was required.

\paragraph{Consent to Participate}Not applicable. This study used publicly available, de-identified retrospective datasets and did not involve direct interaction with human participants.

\paragraph{Consent for Publication}
Not Applicable.

\paragraph{Conflict of Interest}  The authors declare no competing interests.

\bibliography{sn-bibliography}

\end{document}